\documentclass[manuscript,nonacm]{acmart}
\AtBeginDocument{%
  }

\usepackage{graphicx}
\usepackage{float}
\usepackage{booktabs}

\begin{document}

\title{Interaction-Centered Intelligence: Toward an Interaction-Based Theory of Human-AI Co-Creation}

\author{Nicholas Davis}
\email{ndavis35@gatech.edu}
\affiliation{%
  \institution{Co-Creative AI Consulting}
  \city{Elyria}
  \state{Ohio}
  \country{USA}
}

\renewcommand{\shortauthors}{Davis}

\begin{abstract}
Traditional artificial intelligence has largely conceptualized intelligence as isolated computation occurring within bounded agents. Across classical AI, machine learning, and many generative systems, the dominant unit of analysis remains the individual model or autonomous system evaluated through outputs, benchmarks, prediction accuracy, or optimization performance. While these approaches have produced major advances, they often under-theorize the role of interaction in the emergence of intelligence, creativity, meaning, and adaptive behavior. This paper proposes interaction as the primary unit of analysis for co-creative AI and interaction-centered intelligence more broadly. Drawing from distributed cognition, embodied cognition, enaction, participatory sense-making, human-computer interaction, and computational creativity, the paper traces a historical progression toward increasingly relational accounts of intelligence. Building upon prior work in Creative Sense-Making, quantified co-creation, and co-creative systems such as the Drawing Apprentice and AI Drawing Partner, it argues that intelligence emerges through evolving interaction dynamics among agents, environments, and socio-technical systems rather than solely through internal computation. The paper introduces Interaction-Centered Intelligence as a framework for understanding human-AI co-creation, collaborative emergence, adaptive participation, and interactional dynamics. Rather than evaluating intelligence solely through generated outputs, the framework emphasizes interaction trajectories, coordination patterns, participatory engagement, adaptive regulation, and interactional drift unfolding through time. Implications for explainable co-creative AI, hybrid intelligence, enactive AI, and future human-AI systems are discussed.

\end{abstract}




\maketitle

\section{Introduction}
Artificial intelligence research has historically focused on isolated intelligent agents performing tasks through internal computation. Across symbolic artificial intelligence, expert systems, machine learning, and contemporary deep learning architectures, intelligence has often been conceptualized primarily through: representation, prediction, optimization, planning, classification, and autonomous generation \cite{RussellNorvig2021}. Within these paradigms, cognition is generally treated as something occurring inside bounded computational systems, while interaction functions largely as a secondary mechanism for transferring information between otherwise separate entities. Human users become external operators, prompts become inputs, and collaboration becomes reduced to exchanges between independently operating systems. Intelligence is therefore often evaluated primarily through outputs such as: benchmark performance, prediction accuracy, generative capability, or task completion \cite{LeCunBengioHinton2015}.

This output-centered framing has produced major advances in artificial intelligence research. However, many forms of intelligence cannot be adequately explained solely through isolated internal computation. Creativity, improvisation, collaborative adaptation, meaning construction, participatory engagement, and human-AI co-creation emerge dynamically through interaction itself rather than solely inside individual agents \cite{DavisEtAl2016PSM}. In these contexts, intelligence unfolds relationally across evolving systems involving: humans, AI systems, environments, interfaces, materials, representations, and temporal interaction dynamics.

A growing body of work across cognitive science, human-computer interaction, and computational creativity increasingly challenges isolated models of cognition and intelligence. Diverse traditions including distributed cognition, embodiment, enaction, participatory sense-making, and collaborative intelligence have progressively expanded the unit of analysis beyond individual agents toward larger systems of interaction, participation, and coordination. Collectively, these developments suggest that intelligence may often emerge through relationships among agents, environments, technologies, and social systems rather than solely within isolated minds or machines. From this perspective, creativity and intelligence are not reducible to isolated internal processes or final outputs alone, but instead emerge through evolving interaction trajectories unfolding through time.

\begin{figure}[H]
  \centering
  \includegraphics[width=0.9\textwidth]{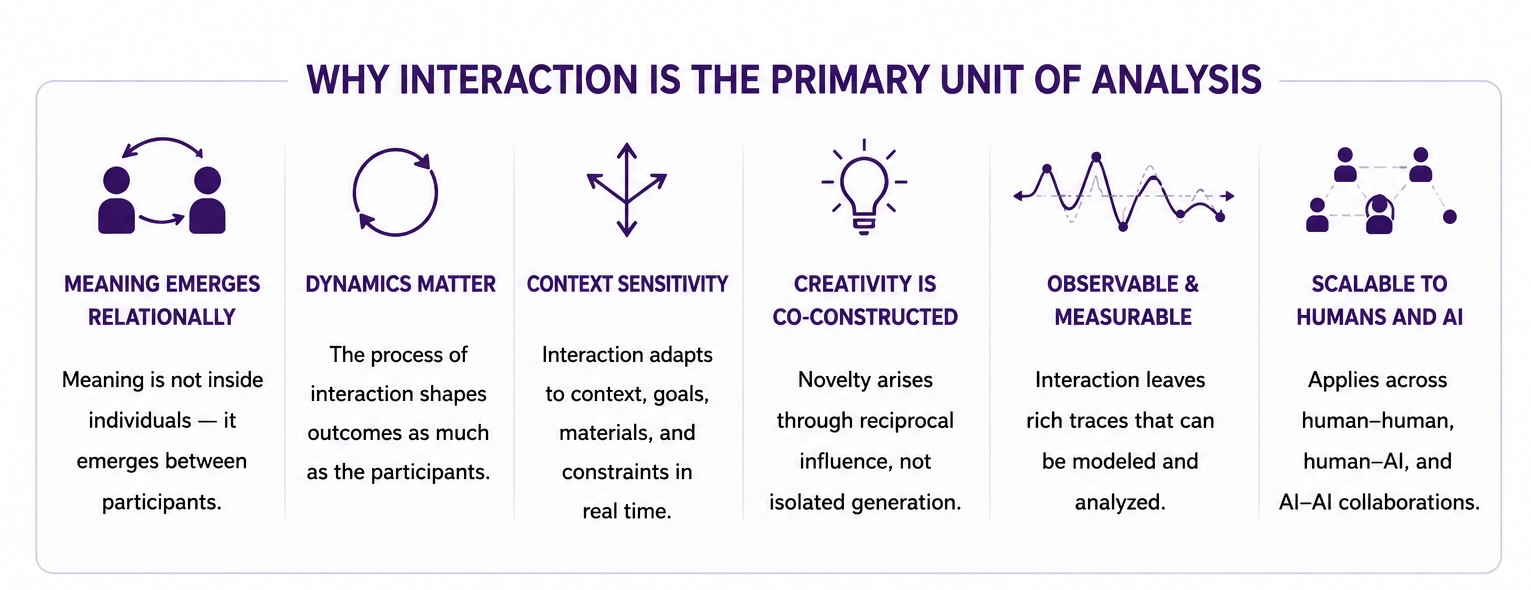}
  \caption{\textbf{Why interaction is a compelling candidate for the primary unit of analysis in co-creative AI.} Traditional AI evaluation focuses on individual agents and outputs, whereas interaction-centered perspectives emphasize the evolving dynamics between participants. The figure illustrates how meaning, adaptation, creativity, and coordination emerge through interaction trajectories that unfold over time and are not fully visible when analyzing either participant in isolation.}
  \label{fig:regCenteredAI}
\end{figure}

\subsection{Research Contributions}

\begin{enumerate}
    \item \textbf{Reframes interaction as the primary unit of analysis in co-creative AI systems.} Rather than treating intelligence as isolated computation occurring solely within humans or machines, the paper argues that creativity, meaning-making, coordination, and adaptive cognition emerge dynamically through interaction trajectories unfolding between humans, AI systems, environments, interfaces, and socio-technical structures. This reframing shifts attention away from static outputs and autonomous generation toward temporally extended participation, collaborative emergence, and evolving interaction dynamics.
    \item \textbf{Synthesizes distributed cognition, embodiment, enaction, participatory sense-making, and co-creative AI into a unified interaction-centered framework.} While prior traditions each expanded the boundaries of cognition beyond isolated symbolic reasoning, this paper argues that these trajectories collectively suggest a broader historical shift toward understanding intelligence as relational, participatory, and dynamically enacted through interaction itself. The paper therefore positions interaction-centered intelligence as a unifying synthesis across previously fragmented theoretical traditions.
    \item \textbf{Positions Creative Sense-Making and quantified co-creation as computational operationalizations of interaction-centered cognition.} Rather than remaining purely philosophical or theoretical, the paper demonstrates how interaction dynamics can be computationally modeled through frameworks involving: activity traces, interaction histories, creative trajectories, participatory rhythms, coordination dynamics, and sense-making curves. These frameworks provide practical approaches for analyzing, visualizing, and quantifying co-creative interaction as it unfolds through time within human-AI systems.
    \item \textbf{Proposes interaction-centered intelligence as a broader paradigm for future human-AI systems.} Rather than viewing AI systems primarily as autonomous generators or isolated problem solvers, interaction-centered intelligence positions future AI systems as dynamically coupled participants within evolving human-AI interaction ecologies. In this sense, the paper establishes interaction-centered intelligence not merely as a framework for co-creative systems, but as a broader research agenda for understanding intelligence itself as an emergent relational phenomenon.
\end{enumerate}

To develop this argument, the paper first examines the limitations of output-centered artificial intelligence frameworks and their difficulty accounting for participatory interaction and collaborative emergence. The paper then traces the historical evolution of the unit of analysis across cognitive science and AI research, beginning with distributed cognition, embodied cognition, enaction, and participatory sense-making. Building upon these foundations, the paper identifies Creative Sense-Making and quantified co-creation as computational frameworks for modeling interaction dynamics during human-AI collaboration. The paper then proposes interaction-centered intelligence as a broader theoretical framework for understanding intelligence as an emergent relational phenomenon unfolding through interaction itself. Finally, the paper discusses implications for human-AI co-creation, hybrid intelligence, explainable AI, and future directions for interaction-centered artificial intelligence systems.

\section{Related Work and Positioning}

Interaction-centered intelligence builds upon and synthesizes multiple prior traditions across cognitive science, human-computer interaction, computational creativity, and human-AI collaboration research. Existing frameworks have increasingly emphasized the importance of interaction, participation, embodiment, and socio-technical systems in understanding cognition and intelligent behavior. However, most existing paradigms still treat interaction primarily as a supporting mechanism, communication channel, interface layer, or contextual factor surrounding otherwise bounded intelligent agents.

\subsection{Human Augmentation and Human--Computer Symbiosis}

The idea that intelligence emerges through collaboration between humans and computational systems has deep roots within the history of computing. Long before contemporary discussions of human--AI collaboration, Licklider (1960) argued for \textit{man-computer symbiosis}, envisioning interactive partnerships in which humans and computers would cooperate in decision making, problem solving, and complex cognitive activity rather than functioning as isolated entities \cite{Licklider1960}. Licklider proposed that humans and computers possess complementary strengths and suggested that future intelligent systems would emerge through increasingly integrated forms of human--computer partnership rather than autonomous machine intelligence alone.

These traditions remain highly influential within contemporary human-centered AI and collaborative intelligence research. However, while augmentation and symbiosis frameworks primarily focus on how computational systems enhance human capabilities, Interaction-Centered Intelligence shifts the explanatory focus toward the interaction itself. The central claim is not merely that humans and machines perform better together, but that intelligence, creativity, and meaning-making emerge through interaction dynamics unfolding through time.
\vspace{0.5em}

\subsection{Situated Action, CSCW, and Embodied Interaction}

A second important lineage emerges from situated-action research, computer-supported cooperative work (CSCW), and embodied interaction. Suchman (1987) challenged plan-based models of action by arguing that intelligent behavior cannot be adequately explained through predefined internal representations detached from context \cite{Suchman1987}. Instead, action is continuously constructed and reconstructed through ongoing interaction with social and material environments. Human activity therefore cannot be fully understood apart from the situations in which it occurs.

Building upon these developments, Dourish (2001) proposed the concept of embodied interaction, arguing that meaning emerges through active participation in the world rather than through abstract information processing alone \cite{Dourish2001}. Embodied interaction emphasized how cognition, interpretation, and coordination arise through situated engagement with people, artifacts, environments, and social practices. This work helped move HCI beyond interface manipulation toward broader questions concerning participation, embodiment, and interaction as fundamental phenomena of human activity.

These traditions are particularly relevant to Interaction-Centered Intelligence because they highlight the limitations of explaining intelligent behavior solely through internal representations, plans, or computational states. Interaction-Centered Intelligence extends this trajectory by proposing that interaction itself should become a primary unit of analysis for understanding intelligence in human--AI systems. Rather than treating interaction as the context surrounding cognition, the framework argues that interaction dynamics constitute a central locus through which meaning, coordination, and adaptive intelligence emerge.
\vspace{0.5em}

\subsection{Human--AI Interaction, Teamwork, and Collaborative Intelligence}

Recent Human--AI Interaction (HAI) and Human-Centered AI (HCAI) research has expanded beyond traditional usability concerns to investigate collaboration between humans and intelligent systems. Contemporary frameworks address issues including transparency, explainability, trust, human oversight, user agency, mixed-initiative interaction, adaptive assistance, and collaborative decision making \cite{Amershi2019,Shneiderman2020,Shneiderman2022}. These approaches increasingly emphasize that successful AI systems must support effective cooperation between humans and machines rather than autonomous performance alone.

One of the most influential developments has been the emergence of Human-Centered AI. Shneiderman's HCAI framework argues that future intelligent systems should maintain high levels of both human control and computational automation, producing systems that are reliable, safe, and trustworthy rather than maximizing autonomy alone \cite{Shneiderman2020}. Within this perspective, AI systems should augment human capabilities, support human responsibility, and preserve meaningful human oversight \cite{Shneiderman2022}. Similar themes appear throughout contemporary HCAI research, where human values, agency, transparency, accountability, and user experience increasingly function as central design principles rather than secondary concerns \cite{XuGeGao2021}.

Building upon these developments, Xu, Ge, and Gao (2021) proposed Human--AI Interaction (HAII) as an emerging interdisciplinary domain specifically focused on understanding and designing interactions between humans and increasingly autonomous intelligent systems \cite{XuGeGao2021}. Their work argues that traditional human-computer interaction frameworks are insufficient for addressing the challenges posed by adaptive AI systems and calls for new theoretical, methodological, and interdisciplinary approaches capable of supporting human-centered AI. More recent HAII research has further expanded this agenda toward human-AI collaboration, joint cognitive systems, shared situation awareness, collaborative intelligence, and human-centered design for intelligent systems \cite{XuGeGao2021}.

A growing empirical literature further examines human--AI teamwork, collaborative intelligence, and hybrid human-machine systems. Recent studies investigate how humans and AI systems coordinate attention, establish shared understanding, allocate responsibilities, communicate uncertainty, and adapt to one another during collaborative activity \cite{Tong2025}. Research on collective attention in human--AI teams demonstrates the importance of shared attentional structures for effective collaboration, while work on mutual theory of mind explores how humans and AI systems develop models of one another's knowledge, intentions, capabilities, and goals during interaction \cite{Zhang2025ToM}. Similarly, field experiments involving AI agents increasingly examine how collaborative performance emerges through dynamic patterns of coordination, communication, and task allocation rather than through isolated system capabilities \cite{JuAral2025}.

Participatory approaches to AI design have increasingly challenged technology-centered development models. The emerging participatory turn in AI emphasizes the direct involvement of stakeholders, communities, domain experts, and end users within AI design, governance, and deployment processes \cite{Delgado2025}. Rather than treating users as passive recipients of intelligent technologies, participatory approaches conceptualize AI development as a collaborative socio-technical process in which values, goals, and system behaviors emerge through ongoing engagement among diverse participants.

\subsection{Computational Creativity and Co-Creative AI}

Within computational creativity and co-creative AI research, substantial work has explored AI systems as collaborative creative partners rather than isolated generative tools \cite{DavisHsiaoPopovaMagerko2015a}. Frameworks such as COFI emphasized the importance of interaction design, communication, and turn-taking within co-creative systems \cite{RezwanaMaher2023}. Recent work on human-AI co-creation, hybrid intelligence, and collaborative design systems similarly explores how humans and AI systems may participate together within shared creative processes \cite{Davis2024CSM}. However, much of the existing co-creative AI literature remains centered on: interaction design taxonomies, collaborative workflows, communication structures, or user perceptions of AI collaborators rather than positioning interaction dynamics themselves as the fundamental unit of intelligence analysis.

Earlier computational creativity research frequently focused on questions surrounding autonomous creative capability, novelty generation, evaluation, and computational models of creativity \cite{Boden2004, ColtonWiggins2012}. Within these paradigms, creative systems were often evaluated according to the originality, value, or quality of generated artifacts. While these approaches significantly advanced computational models of creative production, they often treated creativity as a property of isolated systems rather than an emergent phenomenon arising through collaborative interaction.

The emergence of co-creative AI introduced a substantial conceptual shift by reframing AI systems as collaborative participants capable of engaging users through reciprocal interaction, improvisation, turn-taking, and shared creative activity \cite{YannakakisLiapisAlexopoulos2014, DavisEtAl2015b}. Systems such as the Drawing Apprentice demonstrated how creativity can emerge through ongoing interaction between humans and AI agents rather than solely through autonomous generation. Subsequent frameworks increasingly emphasized collaborative engagement, communication, user experience, and interaction design as central concerns in co-creative systems \cite{GuzdialRiedl2019}.

Recent work has further expanded toward explainable co-creative AI, quantified co-creation, hybrid intelligence, and interaction-centered evaluation frameworks that seek to model how collaboration evolves through time rather than focusing exclusively on final artifacts \cite{DavisEtAl2017CSM,DavisRafner2025}. Nevertheless, many existing approaches still primarily analyze the structure of collaboration rather than treating interaction dynamics themselves as the primary explanatory phenomenon. Interaction-centered intelligence extends this literature by proposing that creativity, meaning-making, and adaptive intelligence emerge through interaction trajectories unfolding between participants rather than residing primarily within either collaborator independently.

\section{Limitations of Output-Centered AI}

Traditional artificial intelligence systems are often evaluated primarily through output-oriented metrics such as benchmark performance, classification accuracy, optimization efficiency, generative capability, or task completion \cite{RussellNorvig2021, Nilsson2009,LeCunBengioHinton2015}. Within these paradigms, intelligence becomes inferred from outputs rather than understood through the evolving interaction processes that produce them. Whether in symbolic AI, machine learning, or modern generative systems, the dominant assumption frequently remains that intelligence resides primarily inside bounded computational agents performing internal computation on external inputs \cite{NewellSimon2007,Brooks1991,LeCunBengioHinton2015}.

This output-centered framing has enabled substantial advances across artificial intelligence research, particularly in areas involving prediction, optimization, pattern recognition, and autonomous generation. Contemporary large language models and generative systems further extend this paradigm by producing increasingly sophisticated text, images, music, and multimedia artifacts through large-scale statistical modeling and generative inference \cite{BrownEtAl2020,BommasaniEtAl2021}. However, several limitations emerge when considering co-creative systems, human-AI collaboration, and interaction-centered forms of intelligence.

\subsection{The Hidden Dynamics of Co-Creation}

Output-centered paradigms frequently obscure the interaction dynamics underlying collaboration itself. In co-creative systems, creativity does not emerge solely from isolated internal generation processes, but through evolving interaction between participants involving reciprocal influence, adaptive coordination, improvisation, turn-taking, collaborative negotiation, and dynamically unfolding creative trajectories \cite{DavisHsiaoPopovaMagerko2015a,YannakakisLiapisAlexopoulos2014}. Final artifacts alone cannot fully capture: how collaboration evolved, how meaning emerged, how participants adapted to one another, or how interaction shaped creative outcomes through time.

As a result, many important dimensions of co-creative interaction remain invisible when evaluation focuses exclusively on outputs rather than interaction dynamics. This limitation becomes especially significant in collaborative creativity systems where: timing, responsiveness, participatory engagement, and adaptive interaction play central roles in the creative process itself \cite{Lubart2005,Fischer2005}.

\subsection{Human Agency in Output-Centered AI Paradigms}

Output-centered paradigms often treat human participation as external to intelligence rather than constitutive of it. Human users become reduced to prompt providers, evaluators, selectors, or consumers of machine-generated outputs rather than active participants within evolving collaborative systems. Interaction becomes framed as a mechanism for controlling or querying intelligent systems rather than as the primary site through which intelligence and meaning emerge relationally. This framing risks under-theorizing the participatory and adaptive nature of human-AI collaboration by positioning AI systems as fundamentally autonomous generators operating independently of human interaction dynamics \cite{DeterdingEtAl2017,Shneiderman2022,XuGeGao2021}.

This limitation becomes increasingly visible in contemporary generative AI systems where human contribution may become obscured within opaque generation pipelines. As generative systems become more capable, questions surrounding: authorship, explainability, participation, collaboration, and hybrid intelligence become increasingly difficult to address through output-oriented frameworks alone \cite{DellermannEtAl2019}.Output-centered approaches struggle to account for temporal interaction dynamics unfolding through time. Creativity, collaboration, and participatory meaning-making are not static events but temporally evolving processes involving: adaptation, coordination, breakdown, recovery, conceptual divergence, and changing collaborative relationships.

\subsection{Collaborative Emergence}

Traditional evaluation paradigms frequently compress these interaction dynamics into static endpoint assessments or isolated artifact evaluations, thereby obscuring the evolving structures of interaction itself \cite{DavisEtAl2017CSM,BeaudouinLafon2004}. This becomes particularly problematic in co-creative systems, hybrid intelligence environments, collaborative design systems, and interaction-centered AI contexts where intelligence emerges through sustained participation and adaptive interaction over time.

More broadly, output-centered paradigms often struggle to account for collaborative emergence — situations in which novel meaning, creativity, or adaptive behavior emerge relationally through interaction rather than being attributable solely to individual participants \cite{davis2013human}. Participatory sense-making frameworks argue that interaction itself can become partially autonomous and constitutive of cognition, meaning that collaborative dynamics cannot always be reduced to isolated internal processes within separate agents \cite{DeJaegherDiPaolo2007}. In co-creative systems, meaning and creativity frequently emerge within the interaction itself rather than solely inside either the human or AI participant independently.

These limitations suggest the need for a broader interaction-centered perspective on intelligence. Rather than evaluating intelligence exclusively through isolated outputs, interaction-centered approaches propose that: interaction trajectories, coordination dynamics, participatory engagement, adaptive regulation, and collaborative emergence should become central phenomena for analysis within co-creative AI systems and human-AI collaboration research more broadly.

\section{From Isolated Cognition to Interaction-Centered Intelligence}
The history of cognitive science and artificial intelligence can be partially understood as a gradual expansion of the unit of analysis used to explain cognition and intelligence. Across the twentieth and early twenty-first centuries, multiple theoretical movements progressively shifted explanations of intelligence away from isolated internal computation toward increasingly relational, embodied, distributed, and interaction-centered frameworks. This progression reflects a broader conceptual transition from understanding cognition as something occurring solely inside bounded minds or computational systems toward understanding intelligence as emerging dynamically through interaction between agents, environments, bodies, artifacts, and social systems.

\subsection{Information Processing: Cognition as Symbol Manipulation}
Early information-processing approaches in cognitive science and symbolic artificial intelligence primarily conceptualized cognition as internal symbolic computation occurring within isolated agents \cite{NewellSimon2007}. Classical cognitive science frequently modeled the mind as an information-processing system analogous to a computer, where cognition involved the manipulation of symbolic representations according to formal computational rules. Similarly, early artificial intelligence systems focused heavily on: logical reasoning, symbolic planning, rule-based inference, search, and internal representation. Within these paradigms, intelligence was largely treated as an internal property of bounded systems, while interaction with environments functioned primarily as input and output channels for otherwise self-contained computational processes \cite{RussellNorvig2021}.

\subsection{Distributed Cognition: Extending Cognition Beyond the Individual}

Distributed cognition expanded the unit of analysis beyond isolated minds by proposing that cognition emerges across coordinated systems involving people, tools, artifacts, representations, and environments rather than solely within individual agents \cite{Hutchins1995}. In contrast to classical information-processing approaches, which treated cognition primarily as internal symbolic computation \cite{NewellSimon2007}, distributed cognition emphasized how intelligent activity is distributed across socio-technical systems.

Hutchins’s studies of navigation teams demonstrated how reasoning and problem solving emerge through interactions among individuals, instruments, representations, and environmental structures rather than residing entirely within any single participant \cite{Hutchins1995}. Tools such as maps, diagrams, interfaces, and representational systems therefore become active components of cognition rather than merely external supports \cite{ClarkChalmers1998}.

This perspective strongly influenced human-computer interaction, CSCW, socio-technical systems research, and creativity support tools by shifting attention toward how cognition and creativity emerge through coordinated interaction among people, technologies, and environments \cite{Shneiderman2007}. Importantly, distributed cognition represented one of the first major theoretical movements to substantially expand the unit of analysis beyond isolated individuals.

\subsection{Embodiment: Cognition Through Bodily Engagement with the Environment}

Embodied cognition further expanded the unit of analysis by emphasizing the role of bodily interaction and sensorimotor coupling in shaping cognition and meaning-making \cite{Clark1997,LakoffJohnson1999,VarelaThompsonRosch1991}. Rather than treating cognition as abstract symbolic computation, embodied approaches argued that intelligent behavior emerges through active engagement between an organism and its environment. Cognition therefore becomes grounded in bodily structure, perception, action, movement, and environmental interaction.

This perspective challenged classical computational theories of mind by proposing that perception and action are fundamentally intertwined. Embodied cognition influenced robotics, ecological psychology, human-computer interaction, and adaptive systems research, where intelligent behavior was increasingly understood as emerging through situated interaction rather than internal representation alone. In robotics, Brooks (1991) demonstrated how adaptive behavior could emerge through direct sensorimotor engagement with environments, while Gibson's affordance theory emphasized that organisms perceive environments in terms of opportunities for action arising through organism-environment relationships \cite{Gibson1979}.

Importantly, embodiment expanded the unit of analysis beyond internal symbolic reasoning toward dynamic organism-environment coupling. However, many embodied approaches remained focused primarily on individual agents interacting with their environments rather than on the coordination dynamics that emerge between multiple participants. While embodiment emphasized relational interaction, less attention was often given to social coordination, collaborative emergence, and participatory interaction.

These limitations became increasingly important in research involving social cognition, collaborative creativity, and human-AI co-creation. Subsequent frameworks such as enaction and participatory sense-making would extend embodiment further by emphasizing how cognition and meaning emerge not only through embodied engagement with environments, but also through evolving interaction dynamics among participants.

\subsection{The Enactive Paradigm: Cognition is Sense-Making}
Enaction extended interaction-centered approaches to cognition by proposing that cognition emerges through dynamic perception-action coupling and adaptive engagement between agents and environments rather than through internal representations of a pre-given world \cite{VarelaThompsonRosch1991}. Rather than conceptualizing cognition as detached symbolic processing, enactive cognition argued that organisms actively enact meaningful worlds through ongoing embodied interaction with their environments. Cognition therefore became understood as: adaptive sense-making, embodied engagement, relational interaction, and ongoing perception-action coupling.

Within enactive frameworks, intelligence emerges through continuous interaction between agents and environments rather than through isolated internal symbolic computation. Enaction emphasized several key concepts including: autonomy, emergence, embodiment, adaptivity, and participatory sense-making. Autonomous systems actively regulate their relationships with environments in ways that sustain viability and continued interaction \cite{DiPaolo2005}. Meaning is therefore not internally stored and later applied to the world; instead, meaning emerges dynamically through participation and embodied engagement itself.

Importantly, enaction shifted the unit of analysis away from isolated internal representations toward ongoing relational interaction itself. Intelligence became conceptualized as something emerging dynamically through adaptive interaction rather than residing statically inside bounded systems.

Participatory sense-making later extended enaction into social interaction by proposing that meaning can emerge relationally through coordination dynamics between participants themselves \cite{DeJaegherDiPaolo2007}. Interaction became partially autonomous and constitutive of cognition rather than merely serving as a communication channel between pre-existing minds. Meaning was no longer treated solely as something internally generated and externally transmitted; instead, collaborative interaction itself became a site of cognitive emergence.

Within participatory sense-making, social interaction is not reducible to isolated cognition occurring independently inside separate individuals. Instead, interaction dynamics themselves contribute to the emergence of: meaning, coordination, collaborative adaptation, participatory engagement, and shared understanding. This represented a major conceptual shift because interaction itself became understood as a site of cognitive emergence rather than merely a transmission mechanism between isolated agents.

However, while enactive cognition and participatory sense-making provided powerful theoretical frameworks for interaction-centered cognition, many enactive approaches remained difficult to operationalize computationally. Concepts such as: sense-making, participatory emergence, adaptive coordination, and relational intelligence were often theoretically rich but challenging to formally model, quantify, or computationally implement within AI systems.

\begin{figure}[h]
  \centering
  \includegraphics[width=\textwidth]{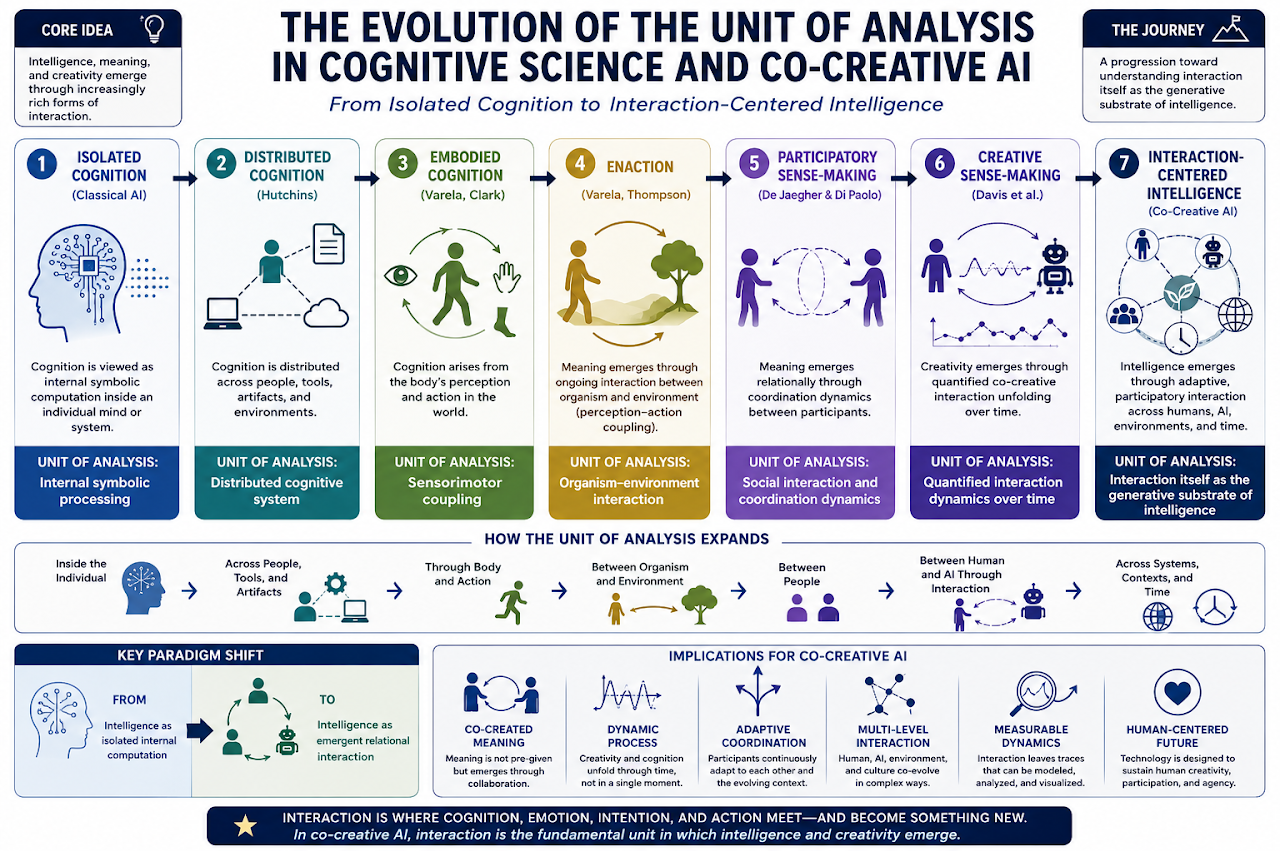}
  \caption{\textbf{Evolution of the unit of analysis from isolated cognition to interaction-centered intelligence.} The figure traces a theoretical progression through information processing, distributed cognition, embodiment, enaction, participatory sense-making, Creative Sense-Making, and interaction-centered intelligence. Across these frameworks, intelligence increasingly shifts from internal computation toward participation, coordination, and interaction unfolding through time.}
  \label{fig:regCenteredAI}
\end{figure}

\subsection{Creative Sense-Making}

Creative Sense-Making (CSM) extended enaction and participatory sense-making into the domain of human-AI co-creation by proposing that interaction itself should become a primary phenomenon for analysis \cite{DavisEtAl2017CSM}. Rather than evaluating creativity solely through generated artifacts or isolated system performance, CSM argued that creativity emerges through evolving interaction among participants during collaborative activity. The framework introduced concepts such as activity traces, creative trajectories, interaction histories, and sense-making curves for modeling how collaboration unfolds through time.

Building upon earlier work in enactive cognition, participatory sense-making, computational creativity, and co-creative AI, CSM reframed co-creation as a temporally unfolding interaction process rather than a sequence of independent contributions. Creativity was understood as emerging through improvisation, reciprocal influence, adaptation, turn-taking, coordination, and collaborative meaning construction between human and AI participants \cite{YannakakisLiapisAlexopoulos2014}.

A central contribution of CSM was the proposal that interaction dynamics can reveal important cognitive and collaborative structures that remain invisible when evaluating final outputs alone. Rather than treating creativity solely as a property of artifacts, the framework conceptualized creativity as an emergent process unfolding through interaction itself. These ideas were explored through studies involving pretend play, collaborative improvisation, and co-creative drawing systems such as the Drawing Apprentice \cite{DavisEtAl2015b,Davis2017Dissertation}.

\begin{figure}[h]
  \centering
\includegraphics[width=0.9\textwidth]{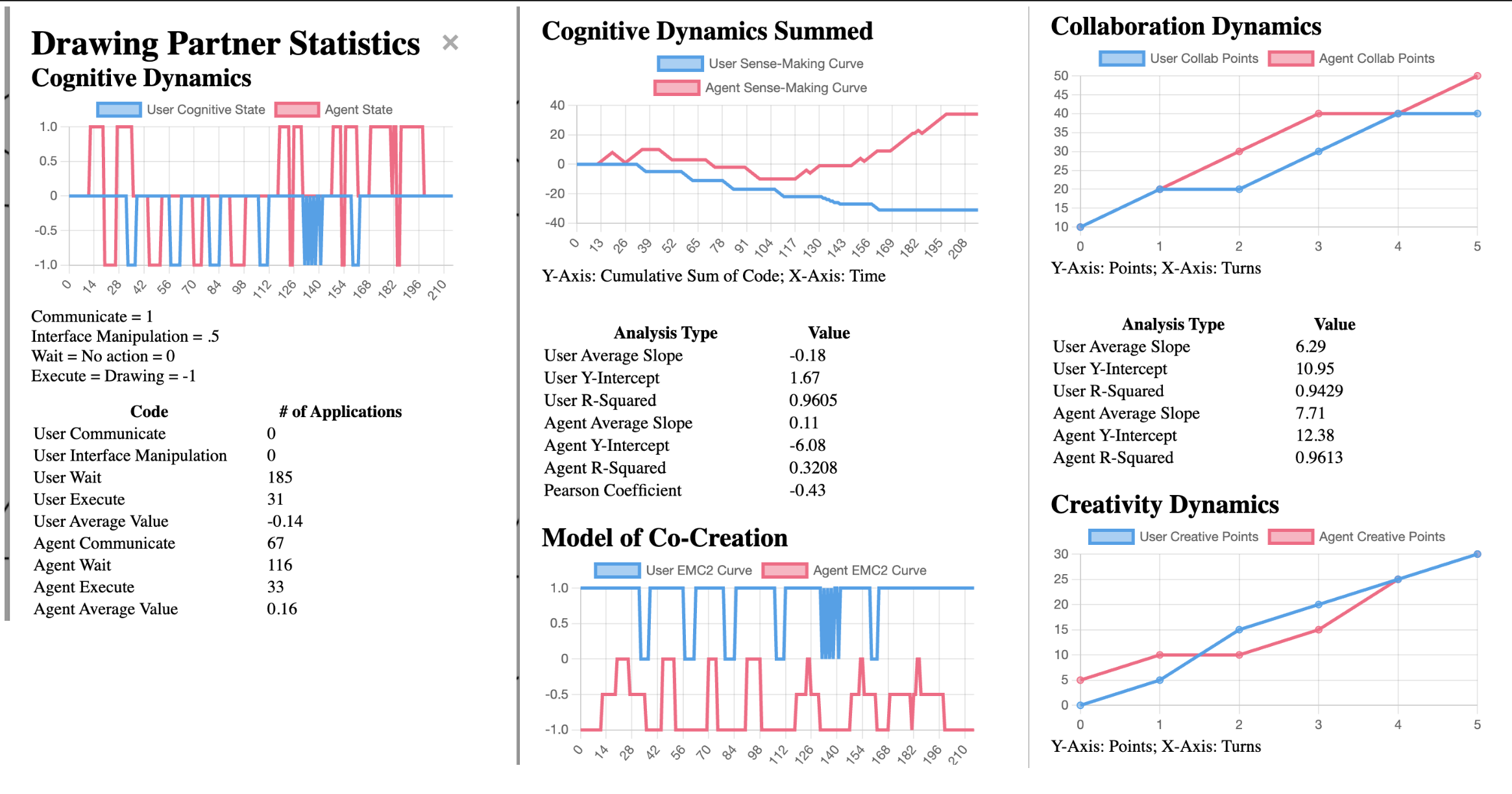}
  \caption{\textbf{Real-time visualization of quantified co-creation within the AI Drawing Partner.} The system continuously models interaction dynamics during collaborative drawing, including cognitive trends, participation balance, collaboration patterns, and creative divergence. These visualizations demonstrate how interaction itself can become observable and measurable during co-creative activity rather than only after collaboration has concluded.}
  \label{fig:IXD1}
\end{figure}

\begin{figure}[h]
  \centering
\includegraphics[width=0.9\textwidth]{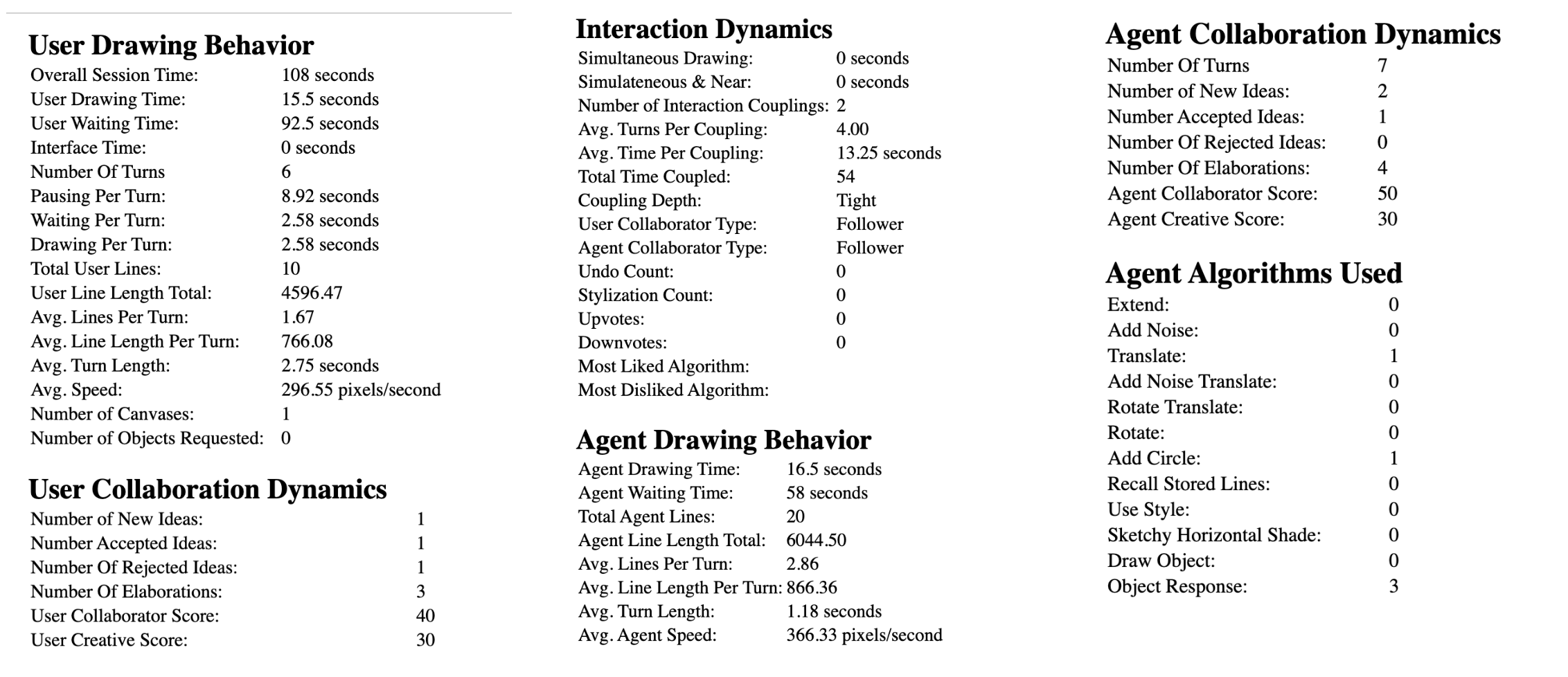}
  \caption{\textbf{Quantified interaction metrics generated during a collaborative drawing session.} The AI Drawing Partner records interaction timing, user and agent behaviors, collaboration dynamics, participation patterns, and system-level interaction statistics. These measures provide a computational representation of co-creative interaction as an evolving process and support interaction-centered evaluation of human–AI collaboration.}
  \label{fig:IXD2}
\end{figure}

By synthesizing enactive cognition with computational interaction modeling, the framework helped operationalize previously difficult-to-model concepts such as: participatory sense-making, collaborative emergence, adaptive coordination, and interaction-centered creativity. More recent work has continued extending the framework into broader models of co-creative AI, quantified collaboration, and interaction-centered intelligence. Davis (2024) further generalized CSM into a broader framework for modeling interaction dynamics in co-creative AI systems, while newer systems such as the AI Drawing Partner operationalized quantified interaction modeling directly inside co-creative systems themselves, as shown in Figure \ref{fig:IXD1} and Figure \ref{fig:IXD2} \cite{DavisRafner2025}.

\subsection{Quantified Co-Creation}

While Creative Sense-Making provided a theoretical framework for understanding how meaning emerges through interaction, Quantified Co-Creation focused on how such dynamics could be observed, measured, and modeled computationally. Rather than evaluating co-creative systems solely through final artifacts, novelty scores, or autonomous generative performance, quantified co-creation proposed that interaction processes themselves should become objects of computational analysis \cite{DavisEtAl2017CSM}. This shift moved attention from what was produced to how collaboration unfolded through time.

Quantified co-creation explored how computational systems could identify and influence evolving creative trajectories. Building upon research in conceptual blending and creative divergence, researchers investigated how AI systems might introduce ambiguity, reinterpretation, or conceptual shifts capable of redirecting collaborative activity toward novel regions of a design space \cite{KarimiDavisGraceMaher2018,KarimiMaherDavisGrace2019}. These efforts extended co-creative AI beyond static generation toward adaptive participation, where systems actively contributed to the evolution of interaction rather than merely producing outputs.

The framework later informed explainable co-creative AI and systems such as the AI Drawing Partner, which integrated interaction analytics directly into the collaborative process \cite{DavisRafner2025}. By capturing and visualizing interaction trajectories, participation dynamics, and collaboration patterns in real time, these systems transformed quantified co-creation from a post-hoc evaluation methodology into an active computational layer for studying and supporting human-AI collaboration. More broadly, quantified co-creation helped establish interaction-centered intelligence as a computationally tractable research direction by providing practical methods for operationalizing and analyzing interaction dynamics within co-creative systems.

\section{Interaction-Centered Intelligence}
Interaction-centered intelligence proposes that intelligence emerges through interaction itself rather than solely through isolated internal computation. Rather than conceptualizing cognition as a property contained entirely within bounded agents, this framework argues that intelligence unfolds dynamically across evolving systems of participation involving: humans, AI systems, environments, interfaces, artifacts, social systems, and temporal interaction dynamics. Within this perspective, intelligence is not reduced to internal symbolic manipulation, prediction, or autonomous generation alone. Instead, cognition emerges relationally through ongoing interaction, adaptive coordination, participatory engagement, and collaborative sense-making between dynamically coupled systems \cite{DeJaegherDiPaolo2007,VarelaThompsonRosch1991}.

\begin{figure}[h]
  \centering
\includegraphics[width=\textwidth]{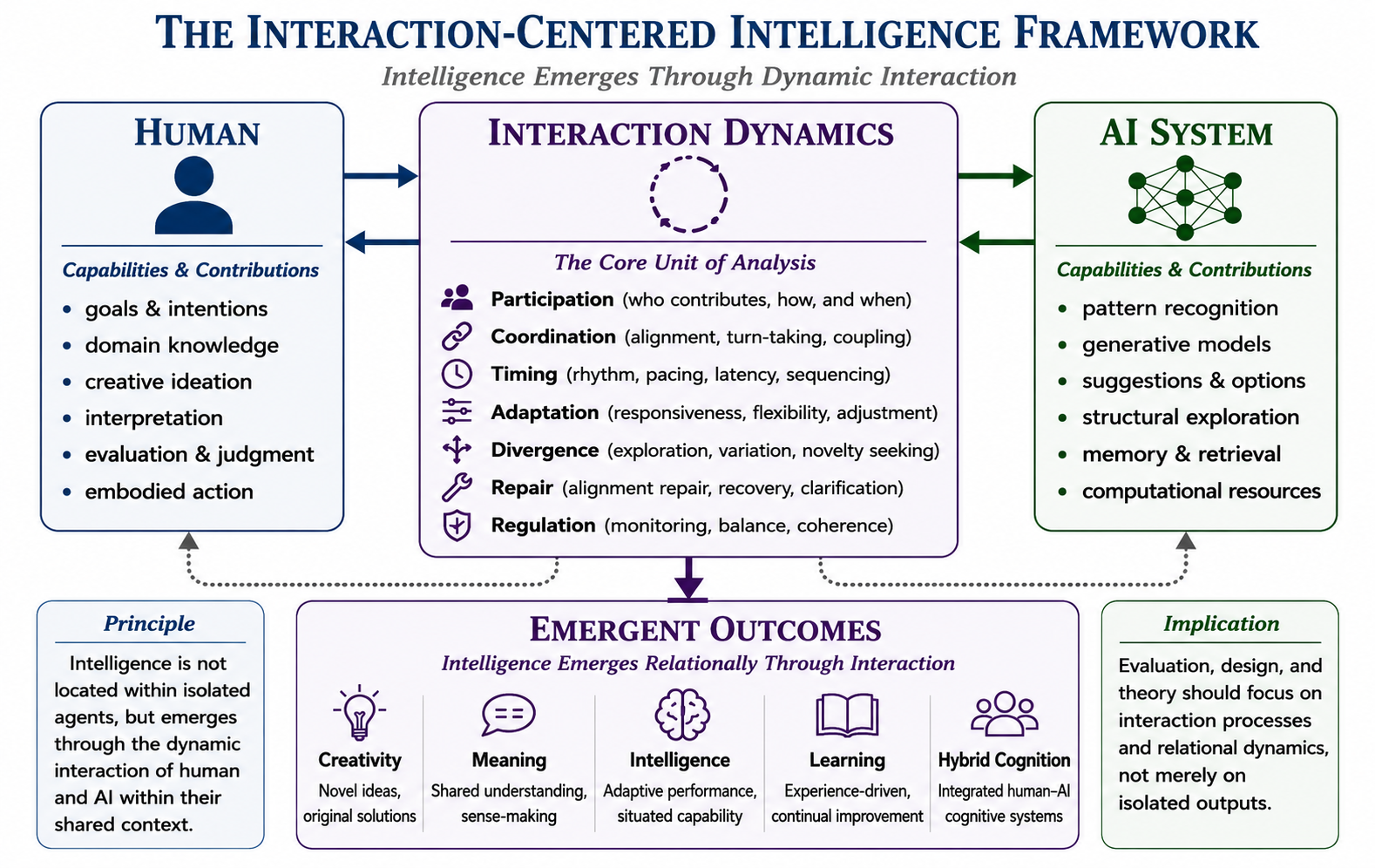}
  \caption{\textbf{The Interaction-Centered Intelligence Framework.} Intelligence emerges through interaction dynamics occurring between humans and AI systems rather than residing exclusively within either participant. Participation, coordination, timing, adaptation, divergence, repair, and regulation form the interactional substrate through which creativity, meaning, learning, hybrid cognition, and intelligent behavior emerge. The framework positions interaction as the primary unit of analysis for co-creative AI and future human–AI systems. This framework reframes intelligence from a property of isolated agents to a property emerging from sustained interaction trajectories among humans, AI systems, and their environments.}

  \label{fig:interactionParadigm}
\end{figure}

This perspective is particularly important in co-creative AI because collaborative intelligence cannot be fully explained through isolated outputs alone. In co-creative systems, creativity emerges through: reciprocal influence, adaptive coordination, improvisation, turn-taking, collaborative negotiation, conceptual divergence, and evolving interaction trajectories \cite{DavisEtAl2017CSM}.

These interaction processes frequently generate forms of meaning and creativity that cannot be attributed solely to either the human or AI participant independently. Instead, intelligence emerges relationally through interaction itself. Interaction-centered intelligence therefore emphasizes: participatory engagement, adaptive regulation, collaborative emergence, coordination dynamics, interaction rhythms, and sustained interaction trajectories as central explanatory phenomena. This perspective reframes co-creative AI systems not merely as creative applications or productivity tools, but as experimental platforms for studying intelligence as an emergent relational phenomenon.

For example, the AI Drawing Partner operationalizes interaction-centered intelligence directly within a co-creative system by modeling interaction dynamics during collaborative drawing sessions through quantified interaction curves, activity traces, and evolving collaboration trajectories \cite{DavisRafner2025}. Rather than functioning solely as an autonomous generator, the system participates dynamically in co-creation while simultaneously modeling how interaction evolves through time.

Interaction-centered intelligence also aligns closely with emerging work in hybrid intelligence and human-centered AI. Hybrid intelligence frameworks increasingly emphasize that future intelligent systems may depend upon effective collaboration between humans and AI systems rather than autonomous machine intelligence alone \cite{DellermannEtAl2019}. Similarly, human-centered AI increasingly focuses on augmenting human capabilities and supporting collaborative interaction rather than replacing human participation entirely \cite{Shneiderman2022}. Interaction-centered intelligence extends these perspectives further by proposing that interaction itself constitutes the generative substrate through which intelligence unfolds.
Importantly, this perspective also reframes intelligence as fundamentally temporal and process-oriented. Intelligence becomes understood not as static capability, but as an evolving interaction trajectory involving: adaptation, coordination, participation, regulation, and collaborative emergence unfolding through time.

From this perspective, intelligence cannot be fully reduced to isolated outputs or internal model states because meaning and adaptive behavior emerge dynamically through interaction itself.

\section{Modeling Interaction Dynamics and Drift in Human-AI Systems}

One of the most important emerging directions within interaction-centered intelligence involves understanding how interaction quality evolves, stabilizes, degrades, or drifts through time during sustained human-AI collaboration. While Creative Sense-Making and quantified co-creation introduced frameworks for modeling interaction dynamics during co-creative activity \cite{DavisEtAl2017CSM}, interactional drift extends these ideas further by focusing specifically on the regulation and maintenance of coherent interaction over extended participatory trajectories.

Traditional AI systems frequently evaluate intelligence through isolated outputs or short-term task performance. However, interaction-centered systems involve temporally extended participation in which collaboration unfolds dynamically through repeated interaction cycles involving: coordination, adaptation, negotiation, pacing, participation balancing, conceptual divergence, and collaborative repair. Over time, these interaction dynamics may gradually stabilize into coherent collaborative structures or drift toward degradation, imbalance, disengagement, or breakdown.

Interactional drift refers to the gradual transformation of interaction quality, coordination structure, participatory balance, or collaborative coherence during sustained interaction between humans and AI systems. Rather than assuming interaction remains stable throughout collaboration, this framework proposes that co-creative systems continuously evolve through dynamic interaction trajectories that may strengthen, destabilize, fragment, or reorganize through time.

This perspective builds upon earlier work in enaction and participatory sense-making, where cognition emerges through ongoing adaptive coupling between agents and environments \cite{VarelaThompsonRosch1991,DeJaegherDiPaolo2007}. Within interaction-centered intelligence, collaboration itself becomes a dynamically regulated system requiring continuous adaptation in order to maintain coherence. Interaction therefore involves not only generation and participation, but also the ongoing regulation of interaction quality itself.

Several forms of interactional drift may emerge within co-creative systems. Participatory imbalance may occur when one participant increasingly dominates interaction while the other becomes passive or disengaged. Temporal drift may involve changes in pacing, responsiveness, rhythm, or coordination timing during collaboration. Conceptual drift may occur when collaborators progressively diverge in meaning construction, goals, or creative direction. Interactional rigidity may emerge when collaborative behavior becomes repetitive, over-stabilized, or insufficiently adaptive. Conversely, excessive divergence may destabilize collaboration entirely by disrupting shared coordination structures.

\begin{table*}[t]
\centering
\caption{Proposed forms of interactional drift within human--AI collaborative systems.}
\label{tab:interactional_drift}

\begin{tabular}{p{3cm} p{5cm} p{5cm}}
\toprule

\textbf{Drift Type} &
\textbf{Description} &
\textbf{Potential Effects} \\

\midrule

Participatory Drift &
Progressive imbalance in contribution,
engagement, or agency between collaborators. &
Reduced collaboration quality, loss of
shared participation, over-reliance on one
participant. \\

Coordination Drift &
Breakdown of synchronization, timing,
or collaborative alignment between participants. &
Miscommunication, interaction friction,
decreased coherence. \\

Conceptual Drift &
Gradual divergence of goals, meanings,
intentions, or creative directions. &
Loss of shared understanding,
fragmented collaboration trajectories. \\

Temporal Drift &
Instability emerging across extended
interaction histories and evolving trajectories. &
Degraded long-term coherence,
interactional inconsistency. \\

Interaction Rigidity &
Reduced adaptability resulting in repetitive,
over-constrained, or inflexible interaction patterns. &
Creative stagnation, reduced exploration,
loss of novelty. \\

Repair Drift &
Failure of collaborative recovery mechanisms
following misunderstandings or breakdowns. &
Escalating interaction failures,
reduced resilience. \\

Coherence Drift &
Accumulated degradation of interactional
stability across multiple dimensions. &
Collapse of collaborative emergence,
reduced system effectiveness. \\

\bottomrule
\end{tabular}

\end{table*}

These forms of drift are particularly important in long-term human-AI systems because sustained collaboration requires ongoing adaptive regulation rather than static interaction patterns. Unlike short-term prompt-response systems, interaction-centered AI environments involve evolving participatory relationships unfolding across extended interaction trajectories. As a result, maintaining coherent collaboration may require systems capable of dynamically regulating interaction itself.

Interactional drift therefore positions adaptive regulation as a central component of interaction-centered intelligence. Future co-creative AI systems may increasingly require mechanisms for: regulating participatory balance,
managing collaborative pacing,
detecting coordination breakdown,
stabilizing interaction rhythms,
repairing collaborative divergence,
and sustaining coherent interaction trajectories over time.

This perspective extends interaction-centered intelligence beyond modeling interaction dynamics alone toward understanding how collaborative systems maintain coherence under conditions of continual adaptation and structural change. Importantly, interactional drift also suggests that intelligence may involve not merely producing successful outputs, but sustaining coherent participation through time within dynamically evolving relational systems. Intelligence therefore becomes partially observable through the ability of collaborative systems to regulate interaction quality, maintain adaptive coordination, recover from breakdowns, and stabilize meaningful participation across extended trajectories of interaction. Interactional drift may therefore represent an important next step beyond Creative Sense-Making and quantified co-creation by extending interaction-centered intelligence toward the study of long-term collaborative stability, adaptive regulation, and evolving participatory coherence within hybrid human-AI systems.

\section{Implications for Human-AI Co-Creation}
Interaction-centered intelligence has several important implications for future human-AI systems, particularly within co-creative AI, hybrid intelligence, human-centered AI, and collaborative interaction design.
First, evaluation frameworks for AI systems should increasingly consider: interaction dynamics, participatory engagement, adaptive coordination, collaboration rhythms, interaction trajectories, and sustained participatory interaction rather than evaluating intelligence solely through isolated outputs or benchmark performance. Traditional AI evaluation paradigms frequently assess systems through: accuracy, optimization metrics, output quality, or autonomous capability.

However, these approaches often fail to capture the collaborative interaction dynamics through which intelligence emerges in co-creative systems \cite{DavisEtAl2017CSM,GuzdialRiedl2019}. In co-creative AI, important dimensions of intelligence involve: responsiveness, improvisation, coordination, communication, and adaptive participation during interaction itself. This shift suggests that future evaluation frameworks may increasingly require longitudinal and interaction-centered methods capable of modeling how collaboration evolves dynamically through time.

Second, interaction-centered intelligence has major implications for explainable AI systems. Traditional explainable AI approaches often focus on exposing: model weights, feature importance, internal representations, or decision pathways.
While these approaches may improve interpretability at the level of model computation, they often under-theorize the collaborative interaction processes through which humans and AI systems jointly construct meaning during co-creation \cite{Miller2019}. Interaction-centered approaches instead suggest that explainability should increasingly involve revealing: evolving interaction dynamics, participatory trajectories, collaborative adaptation, conceptual divergence, and co-regulation processes underlying human-AI collaboration itself.

Recent work in quantified co-creation and explainable co-creative AI increasingly moves in this direction by visualizing interaction curves, activity traces, participation dynamics, and creative trajectories during collaboration \cite{DavisClemensBrowneRezwana2025b}. This reframes explainability not merely as exposing internal model states, but as making collaborative interaction itself visible and interpretable.

Third, co-creative systems may function as important empirical environments for studying interaction-centered cognition itself. Because co-creative AI systems involve ongoing collaboration between humans and AI agents, they provide unique research environments for investigating: participatory sense-making, adaptive coordination, collaborative emergence, interaction dynamics, and interaction-centered intelligence.

\begin{table}[t]
\centering
\caption{Comparison of traditional AI evaluation and interaction-centered evaluation.}
\label{tab:evaluation_comparison}

\begin{tabular}{ll}
\toprule
\textbf{Traditional AI Evaluation} & \textbf{Interaction-Centered Evaluation} \\
\midrule

Accuracy & Coordination quality \\

Task completion & Participation balance \\

Output quality & Interaction trajectory quality \\

Benchmark score & Coherence maintenance \\

Novelty & Collaborative emergence \\

User satisfaction & Adaptive co-regulation \\

Autonomous capability & Hybrid collaborative capability \\

Model explainability & Interaction explainability \\

Static outcomes & Temporal interaction evolution \\

Single-agent performance & Human--AI system performance \\

Optimization efficiency & Participatory engagement \\

Prediction accuracy & Adaptive regulation \\

\bottomrule
\end{tabular}

\end{table}

Systems such as the Drawing Apprentice and AI Drawing Partner demonstrate how co-creative AI platforms can simultaneously function as: collaborative creative systems, cognitive research platforms, and interaction modeling environments \cite{DavisRafner2025}. This suggests that co-creative systems may play an increasingly important role in operationalizing and studying interaction-centered cognition computationally.

\subsection{Interaction-Centered Intelligence Testable Hypotheses}

A central claim of interaction-centered intelligence is that intelligence, creativity, and meaning emerge through evolving interaction dynamics rather than solely through isolated internal computation or generated outputs. While this paper has primarily developed this argument conceptually an important implication of the framework is that it generates empirically testable hypotheses about human-AI collaboration. If interaction constitutes a primary explanatory substrate for intelligence, then measurable properties of interaction should predict important collaborative outcomes. Interaction-centered intelligence therefore shifts scientific attention toward the analysis of interaction trajectories, participatory dynamics, coordination structures, adaptive regulation, and collaborative emergence as observable phenomena capable of empirical investigation.

Several initial hypotheses emerge directly from this framework.

\textbf{H1: Interaction coherence predicts co-creative success.}

Collaborative systems exhibiting higher levels of interactional coherence—including stable coordination, adaptive participation, responsive turn-taking, and sustained collaborative alignment—should produce more successful co-creative outcomes than systems exhibiting fragmented or unstable interaction dynamics. Success may be measured through user evaluations, collaborative performance, perceived creativity, or longitudinal engagement.

\textbf{H2: Interactional drift predicts collaboration breakdown.}

Accumulating forms of interactional drift, including participatory imbalance, coordination drift, conceptual divergence, temporal instability, and coherence degradation, should predict reductions in collaborative effectiveness and increased likelihood of interaction breakdown. Interaction-centered approaches therefore predict that monitoring interaction trajectories may provide earlier indicators of collaboration failure than output evaluation alone.

\textbf{H3: Participation balance predicts perceived collaboration quality.}

Co-creative systems exhibiting more balanced patterns of participation between human and AI collaborators should be associated with higher perceptions of agency, partnership, engagement, and collaborative satisfaction. Excessive dominance by either participant may reduce perceived co-creativity by weakening participatory contribution and collaborative emergence.

\textbf{H4: Adaptive repair predicts long-term human-AI engagement.}

The ability of human-AI systems to recover from misunderstandings, coordination failures, conceptual divergence, and interactional disruptions should positively predict sustained engagement over extended interaction trajectories. Interaction-centered intelligence predicts that successful collaboration depends not only on generation quality, but on the capacity for ongoing adaptive repair and re-coordination.

\textbf{H5: Interaction dynamics predict creative outcomes beyond output metrics alone.}

Measures derived from interaction histories—including participation rhythms, activity traces, coordination patterns, interaction trajectories, and sense-making curves—should explain variance in creative outcomes beyond traditional output-centered measures such as artifact quality or task completion. This hypothesis follows directly from Creative Sense-Making and quantified co-creation frameworks, which propose that important dimensions of creativity emerge within interaction itself rather than solely within final artifacts.

\textbf{H6: Collaborative emergence exhibits measurable interaction signatures.}

Periods of perceived insight, novelty, creative breakthrough, or shared meaning construction should correspond to identifiable interaction patterns involving adaptive coordination, reciprocal influence, conceptual divergence, and subsequent stabilization. If collaborative emergence is a genuine interactional phenomenon, it should become observable through recurring structures within quantified interaction traces.

Together, these hypotheses suggest that interaction-centered intelligence should be understood not merely as a conceptual reframing of cognition and AI, but as an emerging empirical research program. Frameworks such as Creative Sense-Making, quantified co-creation, activity traces, interaction trajectories, and interactional drift provide initial computational methods for investigating these hypotheses within co-creative AI systems. Future research may therefore evaluate interaction-centered intelligence not solely through theoretical argumentation, but through systematic empirical study of how interaction dynamics contribute to the emergence, maintenance, and regulation of intelligence within human-AI collaborative systems.

\section{Ethical Considerations and Participatory Governance}

Interaction-Centered Intelligence carries important ethical implications for the design, evaluation, and governance of future human--AI systems. Traditional discussions of AI ethics frequently focus on properties of models themselves, including fairness, bias, transparency, privacy, accountability, safety, and robustness. While these concerns remain critically important, an interaction-centered perspective suggests that ethical evaluation cannot be limited to the internal properties of AI systems alone. If intelligence emerges through interaction, then ethical responsibility must also be understood at the level of interactional systems and the socio-technical contexts within which they operate.

Recent Human-Centered AI research emphasizes that intelligent systems should remain reliable, safe, and trustworthy while preserving meaningful human control and agency \cite{Shneiderman2020,Shneiderman2022}. From this perspective, successful AI systems are not those that maximize autonomy at the expense of human participation, but those that support human capabilities while maintaining transparency, accountability, and user oversight. Interaction-Centered Intelligence aligns strongly with these principles. However, it extends them by arguing that ethical evaluation should include not only how an AI system behaves, but also how participation, coordination, and influence are distributed across an evolving human--AI interaction.

This perspective becomes particularly important in the context of participatory AI design. Recent work on the participatory turn in AI argues that stakeholders, communities, and end users should play active roles in shaping the goals, values, behaviors, and governance structures of intelligent systems \cite{Delgado2025}. If interaction constitutes the primary locus through which intelligence emerges, then ethical questions must include who shapes those interactions, whose values become embedded within them, whose contributions are recognized, and who possesses the authority to modify or regulate them. Interaction-centered systems therefore require attention not only to technical performance, but also to participation, representation, inclusion, and power within socio-technical ecosystems.

Interaction-centered perspectives also highlight the ethical significance of human labor within AI systems. Contemporary AI often obscures the extensive human contributions involved in data generation, annotation, evaluation, prompting, coordination, oversight, and collaborative use. In co-creative and collaborative systems, meaningful outcomes frequently emerge through joint participation rather than autonomous machine activity alone. Ethical frameworks should therefore recognize the distributed nature of authorship, contribution, and responsibility within human--AI interaction. Questions of attribution, ownership, credit, and creative agency become increasingly important when intelligence and creativity emerge relationally through collaboration rather than residing solely within either participant.

Many of these factors emerge not only from model outputs but also from patterns of human-AI interaction, including over-reliance, loss of agency, automation bias, misinformation propagation, unequal participation, and failures of oversight. Interaction-Centered Intelligence therefore supports governance approaches that evaluate how risks emerge within ongoing socio-technical interactions rather than exclusively within isolated computational systems.

Meaningful human control represents another central ethical requirement. As AI systems become increasingly proactive, adaptive, and collaborative, maintaining user agency becomes more challenging. Human-centered frameworks consistently argue that intelligent systems should augment rather than replace human judgment and participation \cite{Shneiderman2020}. Interaction-Centered Intelligence similarly argues that healthy human--AI systems should support adaptive participation rather than passive dependence. Ethical interaction design should therefore promote transparency, contestability, reversibility, and opportunities for human intervention throughout the interaction process.

From this perspective, ethics becomes inseparable from interaction. The central ethical challenge is not merely building better models, but cultivating interactional systems that support human flourishing, preserve meaningful agency, distribute participation fairly, manage risks responsibly, and enable collaborative forms of intelligence that remain accountable to the people and communities they affect.

\section{Limitations and Challenges of Interaction-Centered Intelligence}

While interaction-centered intelligence offers a promising framework for understanding co-creative AI systems and human-AI collaboration, several limitations and open challenges remain.

First, interaction-centered intelligence remains partially theoretical and conceptual in its current form. Although frameworks such as Creative Sense-Making and quantified co-creation provide computational approaches for modeling interaction dynamics, many aspects of interaction-centered cognition remain difficult to operationalize consistently across domains. Questions remain regarding what precisely constitutes an interaction, how interaction boundaries should be defined, and which temporal scales are most appropriate for analysis. Different domains may require substantially different interaction metrics, modeling approaches, and representational frameworks \cite{BeaudouinLafonBodkerMackay2021}. 

Second, measuring interaction dynamics remains inherently challenging because interaction unfolds across multiple simultaneous layers, including communication, coordination, timing, adaptation, social context, perception, and environmental coupling. While quantified co-creation introduces approaches involving activity traces, trajectories, and sense-making curves, these metrics do not yet capture the full richness of participatory interaction. Interaction-centered intelligence therefore should not be interpreted as a claim that all meaningful interaction dynamics are fully quantifiable. Instead, current approaches represent partial computational approximations of much more complex relational processes.

Third, operationalizing interaction-centered intelligence computationally at scale remains an open research challenge. Current generative AI systems still largely rely on prompt-response architectures and output-centered evaluation paradigms. While interaction-centered frameworks emphasize sustained participatory engagement, many existing AI infrastructures are not yet designed to model long-term collaborative trajectories, adaptive co-regulation, or evolving interaction histories across time.

Despite these limitations, interaction-centered intelligence provides a useful conceptual and computational direction for future human-AI systems. Rather than replacing existing approaches entirely, the framework is intended to complement and extend prior work involving distributed cognition, embodiment, enaction, participatory interaction, and human-centered AI. The framework therefore should be understood as an evolving research agenda rather than a finalized theory of intelligence.

\section{Future Directions}

Interaction-centered intelligence suggests several promising directions for future research. First, new evaluation frameworks are needed that move beyond benchmark performance and output quality toward the analysis of interaction dynamics themselves. Future work should investigate computational methods for measuring participation, coordination, interaction trajectories, adaptive responsiveness, coherence maintenance, and collaborative emergence across human-AI systems. Such approaches may provide a richer understanding of intelligence than evaluation based solely on isolated outputs.

A second direction involves the development of adaptive regulation mechanisms capable of maintaining coherent collaboration over extended interaction histories. Interaction-centered intelligence suggests that successful human-AI systems may depend not only on generation quality, but also on the ability to regulate participatory balance, manage interaction pacing, detect coordination breakdowns, support collaborative repair, and adapt to interactional drift. Understanding how collaborative systems stabilize, reorganize, and sustain meaningful participation through time remains an important open challenge.

Finally, future work should extend interaction-centered intelligence beyond co-creative AI into broader domains including education, scientific discovery, decision support, organizational collaboration, human-robot interaction, and hybrid intelligence systems. As AI becomes increasingly embedded within human cognitive and social environments, understanding how intelligence emerges through interaction may become essential for designing future human-centered AI systems. More broadly, interaction-centered intelligence suggests a research agenda focused not only on what intelligent systems produce, but on how intelligence emerges, evolves, and is sustained through participation itself.

\section{Conclusion}

This paper argued that interaction should become the primary unit of analysis in co-creative AI systems and interaction-centered intelligence more broadly. While traditional artificial intelligence has largely conceptualized intelligence as internal computation occurring within isolated agents, a growing body of work spanning distributed cognition, embodied cognition, enaction, participatory sense-making, computational creativity, and human-AI collaboration increasingly suggests that intelligence often emerges through relationships among agents, environments, artifacts, and interaction processes rather than within any single system alone. Building upon these traditions, Interaction-Centered Intelligence proposes that creativity, meaning-making, coordination, adaptation, and collaborative intelligence emerge through evolving interaction dynamics unfolding across humans, AI systems, and their shared environments. From this perspective, co-creative AI systems are not merely applications that produce creative outputs, but empirical environments for studying how intelligence emerges through participation, coordination, and adaptive interaction. As AI becomes increasingly integrated into human creative, cognitive, educational, and social activities, understanding interaction itself may become as important as understanding the individual systems that participate within it. Rather than asking only what intelligent systems can produce, Interaction-Centered Intelligence asks how intelligence emerges through interaction—and proposes that this interactional level may provide a critical foundation for understanding future human-AI systems.

\section{Acknowledgments}

The author would like to acknowledge the use of ChatGPT (OpenAI) as a collaborative research and writing assistant during the development of this manuscript. Through an extended iterative dialogue, the system contributed to brainstorming, theoretical refinement, literature organization, conceptual critique, editorial feedback, figure development, and manuscript revision. The ideas, arguments, interpretations, and conclusions presented in this paper remain the sole responsibility of the author.

\bibliographystyle{ACM-Reference-Format}
\bibliography{pubsRevise2}


\end{document}